\documentclass[]{infiniai}
\usepackage{microtype}
\usepackage{graphicx}
\usepackage{booktabs}
\usepackage{array}
\usepackage{tabularx}
\usepackage{amsmath}
\usepackage{subcaption}
\usepackage{textcomp}
\usepackage{wrapfig}
\usepackage{xurl}          
\usepackage{fontawesome5}     
\DeclareRobustCommand{\linkicon}[2]{\makebox[1.5em][l]{#1}#2}  

\usepackage{caption}
\graphicspath{{figures/}}
\colorlet{RefFill}{infiniline!45}
\colorlet{WinnerFill}{infinitint}

\makeatletter
\newcommand{\keepspace}[1]{\par\begingroup\dimen@=\dimexpr\pagegoal-\pagetotal\relax\ifdim\dimen@<#1\newpage\fi\endgroup}
\makeatother

\usepackage[capitalize,noabbrev]{cleveref}

\title{CompressionBench: How can your coding agents compress LLMs?}

\author{Xinrui~Zhong$^*$, Zhifeng~Li$^*$, Zhiming~Wang, Xinze Feng, Yongqi~Liu, Shanglin~Yang,\\ Yang~Zhou, Zikun~Li, Zhuoming~Chen, Beidi~Chen}
\contact{\{xinruizh, zhifengl, zhimingw, xinzef, yongqil, shangliy, yangzho6, zikunl, zhuominc, beidic\}@andrew.cmu.edu}
\affiliation{Carnegie Mellon University}
\contribution[*]{Equal contribution.}
\metadata[\linkicon{\faGithub}{Code}]{\href{https://github.com/Infini-AI-Lab/Compression-Bench}{\texttt{github.com/Infini-AI-Lab/Compression-Bench}}}
\metadata[\linkicon{\faBlog}{Blog}]{\href{https://infini-ai-lab.github.io/Compression-Bench/}{\texttt{infini-ai-lab.github.io/Compression-Bench}}}

\abstract{Recursive self-improvement requires AI systems to improve the models
and infrastructure used by subsequent systems. LLM compression offers
a concrete efficiency objective: retain a pretrained model's behavior
within a smaller model storage budget. We study whether coding agents can
carry out this step autonomously. We formulate compression as a terminal
task in which an agent receives a source checkpoint, storage limit,
inference environment, and time budget, then constructs compressed
weights and executable inference code. The agent controls its algorithm
and experiments; perplexity feedback guides search, and independent
verification checks the final model. Across six agents and six
model--size settings, 29 of 36 submissions pass verification. Selecting
the lowest-perplexity agent design per task yields an average 3.4\%
perplexity increase over BF16 and 77.33\% downstream accuracy, compared
with 79.56\% for BF16. Recorded experiments reveal how agents achieve
these outcomes and where improvement breaks down. Agents lower PPL in 21 of 24 revisions at nearly fixed storage;
complete-model feedback reverses 17 of 79 prior local rankings;
and generation and inference tests expose failures that likelihood alone
misses. The results identify representation design, experimental
judgment, and capability validation as central to agent-driven model
compression, a measurable component of efficiency-oriented
self-improvement.
}

\hypersetup{
  pdftitle={CompressionBench: How can your coding agents compress LLMs?},
  pdfauthor={Xinrui Zhong; Zhifeng Li; Zhiming Wang; Xinze Feng; Yongqi Liu; Shanglin Yang; Yang Zhou; Zikun Li; Zhuoming Chen; Beidi Chen}
}

\begin{document}

\maketitle

\section{Introduction}
\label{sec:introduction}
\begin{figure}[!h]
\centering
\includegraphics[width=0.9\textwidth]{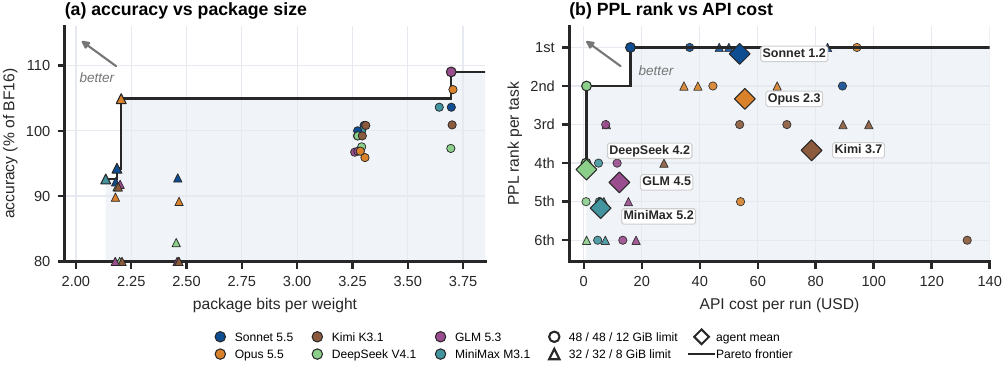}
\caption{Agent-driven compression outcomes.
(a) Mean GSM8K, HumanEval, and TriviaQA accuracy relative to BF16 versus
package bits per original parameter; values below 80\% lie on the axis.
(b) API cost versus within-task PPL rank; diamonds average six tasks per
agent. Pareto frontiers use passing models for the accuracy--size
\mbox{comparison in (a).}}
\label{fig:tradeoff-teaser}
\label{fig:agentic-cost}
\end{figure}

Large language model (LLM) inference is increasingly bounded by memory
traffic rather than arithmetic: in autoregressive decoding, every
generated token reads the model weights from GPU memory, and an off-chip
memory access costs far more energy than the arithmetic it feeds
\citep{horowitz2014computing}. Compression that preserves a model's
behavior within a smaller storage budget therefore reduces both the
memory a deployment occupies and the energy spent on every token it
generates, making models cheaper to serve and runnable on smaller
hardware \citep{frantar2023gptq,lin2024awq,lin2025qserve}. Constructing
such a model remains substantial experimental and systems work, the
kind of work that coding agents increasingly undertake
\citep{novikov2025alphaevolve,zhang2026darwin}. An agent that compresses
models would improve the efficiency of the AI systems that follow it, a
concrete step toward recursive self-improvement. We therefore ask:
\emph{can coding agents autonomously construct compressed LLMs that
remain useful within a given storage budget?}

\begin{figure*}[t]
\centering
\includegraphics[width=\textwidth]{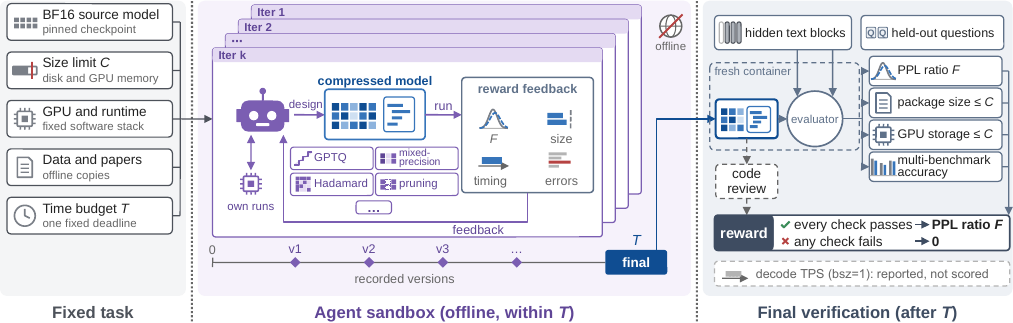}
\caption{Compression through terminal interaction. A task fixes the source
checkpoint, storage allowance $C$, GPU and runtime, and search budget $T$.
The agent constructs and tests compressed models, using reward feedback to
revise weights and inference code. At the deadline, a separate evaluator
checks the frozen artifact's likelihood, generated answers, storage, and
execution. Decode speed (TPS, batch size one) is reported separately from reward.}
\label{fig:overview}
\end{figure*}

Existing compression methods provide a rich set of building blocks,
but no single method yields a useful model under an arbitrary storage
limit and hardware target. To construct one autonomously, an agent
needs three abilities:
\begin{itemize}
\item \textbf{Coordinating coupled numerical decisions.} A storage
budget ties together representation, precision allocation, and
parameter retention: bits spent on one component are unavailable to
the rest, and pruning releases bytes but removes capacity. Even at the
same nominal precision, clipping, rounding, calibration, and metadata
change the quality of a representation
\citep{frantar2023gptq,shao2024omniquant}. The agent must judge
feasible combinations rather than optimize each choice independently.
\item \textbf{Interpreting incomplete feedback.} Weight reconstruction
error, layer-output error, complete-model likelihood, and generated-answer
accuracy each answer a different question. A change that improves a
local approximation can damage the complete model, and a small
likelihood gain can conceal a generation regression. The agent must
decide when inexpensive local evidence suffices and when a
complete-model experiment is worth its cost.
\item \textbf{Co-designing efficient execution.} Compression saves
memory and energy only when unpacking, reconstruction, and indexing do
not erase the reduction in memory traffic, and persistent dequantized
copies can erase the resident-storage savings of a small checkpoint.
The agent must implement an inference path suited to the target
hardware, as in quantized inference systems
\citep{lin2024awq,lin2025qserve}, and verify that it runs correctly.
\end{itemize}

To evaluate these abilities without prescribing how agents compress,
we formulate \textbf{LLM compression as a terminal task}
(\Cref{fig:overview}), following the outcome-based evaluation of
Terminal-Bench and Harbor \citep{merrill2026terminalbench,harbor2026}.
A coding agent reads papers, writes code, constructs checkpoints, and
runs experiments within a fixed time budget. Three designs support this
evaluation.
\textbf{1.~Task: a complete model under a storage limit.} Given source
weights, a storage limit, a GPU, and a pinned runtime, the agent may
combine quantization, mixed precision, pruning, low-rank
representations, and distillation, and must implement the inference
operations its representation requires (\Cref{sec:task}).
\textbf{2.~Feedback: complete-model likelihood.} A supplied evaluator
returns the source-to-compressed perplexity (PPL) ratio on
continuations sampled from the source model; the expected excess
cross-entropy equals the next-token KL divergence from the source
(\Cref{sec:reward}). Evaluations share the agent's time budget, so the
agent decides when local evidence suffices and when a complete-model
measurement is worth its cost.
\textbf{3.~Validation: the executable artifact.} Independent
verification runs the submitted weights through the submitted inference
code on held-out contexts. Both the saved package and the resident GPU
weights must fit the limit; GSM8K, HumanEval, and TriviaQA accuracy must
retain 80\% of BF16; and review checks integrity. Decode speed is
reported separately (\Cref{sec:verification}). Only verified models
receive a score.

In \Cref{sec:evaluation}, we evaluate the models that six coding agents
deliver for three source models with two storage limits each. Of the 36 resulting models, 29 pass verification, including
all 18 produced under the looser limits. Selecting the lowest-PPL agent
design per task yields an average 3.4\% PPL increase over BF16 and
77.33\% downstream accuracy, compared with 79.56\% for BF16
(\Cref{sec:main-results}). In \Cref{sec:analysis}, we analyze the
agents' recorded searches to explain how they improve compression and
where improvement breaks down, yielding three findings.
\emph{Representation design creates room at fixed storage.} In 21 of 24
revisions that change package size by at most 0.01~GiB, agents lower
PPL by improving the quantizer or redistributing precision
(\Cref{sec:numerical-design}).
\emph{Complete-model feedback changes selection.} Reward feedback
reverses 17 of 79 prior local rankings, leading agents to revert
candidates or revise their local validation
(\Cref{sec:validation-design}).
\emph{Likelihood alone does not validate a model.} Intermediate models
gain and lose correct answers at nearly unchanged PPL, and consistency
tests expose inference bugs that likelihood misses
(\Cref{sec:capacity-precision}).

\section{Compression as a Terminal Task}
\label{sec:benchmark}
A compression task asks an agent to produce a model that approximates
a source LLM under a specified storage allowance. Its output includes
both the numerical representation and the code that runs it. The task couples
model approximation with implementation and experimental search.

\subsection{Task Specification and Search Space}
\label{sec:task}
A task specifies a source model $p$ with its tokenizer, a storage limit
$C$, a hardware allocation, a pinned inference runtime, and a wall-clock
budget $T$. The agent works in an offline container with read-only source
weights, starter inference code, searchable compression papers,
calibration and recovery data, and a separate set of questions for local
testing. The runtime provides scheduling and uncompressed operations.
The agent chooses its compression algorithm, produces the weights, and
implements the operations required by its chosen representation.

Let $a=(W,I)$ denote a candidate, where $W$ contains compressed weights
and associated metadata and $I$ is its inference implementation.
Together they define the executable model $q_a$. The search space
includes quantization formats and rounding rules, component-specific
precision, pruning, low-rank representations, and distillation. Changes
to this representation can also require changes to $I$. The task does
not prescribe an order for making these decisions.

The storage allowance constrains two quantities. $S_{\mathrm{pkg}}(a)$
counts the saved model package, including code and tokenizer assets.
$S_{\mathrm{res}}(a)$ counts model storage retained on the GPU during
inference, including scales, codebooks, padding, and persistent
dequantized weights. Shared allocations count once. KV caches, recurrent
states, and temporary computation workspace are excluded. These rules
prevent a small saved checkpoint from satisfying the task by retaining
a full-precision model during execution. The allowance constrains resident weights rather than total peak
GPU~memory.

\subsection{Experiments and Feedback}
\label{sec:agentic-design-loop}
The agent allocates $T$ among implementation, calibration, recovery,
local tests, and evaluation of complete candidates. It can use local
measurements such as reconstruction error, KL divergence, or negative
log-likelihood to choose what to build. A supplied evaluator returns
PPL-based reward, package and resident model storage, decode timing,
and execution diagnostics. All such calls consume the same wall-clock
budget as model construction. The agent must judge whether another evaluation is worth its cost.

This interaction follows the budgeted experimentation setting of
MLAgentBench and RE-Bench
\citep{huang2024mlagentbench,wijk2025rebench}. Candidate versions and
experiments are recorded so that the final model can be connected to
the revisions that produced it. At time $T$, the submitted candidate is
frozen. A separate container performs final verification, adding
held-out generated-answer tests and implementation review to the
likelihood and storage measurements.

\subsection{Likelihood Objective}
\label{sec:reward}
Let $\mathcal D_{\mathrm{s}}$ and $\mathcal D_{\mathrm{v}}$ denote the
search and verification datasets. Both contain continuations sampled
from the source model, using disjoint contexts. A dataset
$\mathcal D=\{(h_n,y_n)\}_{n=1}^{N}$ records each scored token $y_n$
and its context and preceding continuation $h_n$. Source and compressed
models score the same $N$ continuation positions. Their empirical
cross-entropy and perplexity are
\begin{equation}
\begin{aligned}
\widehat H_{\mathcal D}(q)
  &= -\frac{1}{N}\sum_{n=1}^{N}\log q(y_n\mid h_n),\\
\operatorname{PPL}_{\mathcal D}(q)
  &= \exp\!\bigl(\widehat H_{\mathcal D}(q)\bigr).
\end{aligned}
\label{eq:ppl}
\end{equation}
For candidate $a=(W,I)$ with executable model $q_a$, the empirical
excess cross-entropy is
\begin{equation}
\widehat\Delta_{\mathcal D}(a)
  = \widehat H_{\mathcal D}(q_a)-\widehat H_{\mathcal D}(p).
\label{eq:excess-loss}
\end{equation}
Under source sampling, its expectation equals the next-token KL
divergence averaged over scored histories:
\begin{equation}
\mathbb E\bigl[\widehat\Delta_{\mathcal D}(a)\bigr]
 = \mathbb E_{h\sim\mu_p}\!\left[
 D_{\mathrm{KL}}\!\left(p(\cdot\mid h)\,\Vert\,q_a(\cdot\mid h)\right)
 \right].
\label{eq:expected-kl}
\end{equation}
Here $\mu_p$ is the distribution of scored histories induced by the
context mixture and source continuations. This objective measures likelihood preservation on source histories; answer tests
assess the candidate's autoregressive~behavior.

The task minimizes the search estimate under separate saved and resident
model-storage constraints:
\begin{equation}
\begin{aligned}
\underset{a=(W,I)}{\operatorname{minimize}}\quad
  &\widehat\Delta_{\mathcal D_{\mathrm{s}}}(a)\\
\operatorname{subject\ to}\quad
  &S_{\mathrm{pkg}}(a)\leq C,\\
  &S_{\mathrm{res}}(a)\leq C.
\end{aligned}
\label{eq:task-objective}
\end{equation}
Construction, local tests, and supplied evaluations share the wall-clock
budget $T$. The evaluator returns the source-to-compressed PPL~ratio:
\begin{equation}
\begin{aligned}
F_{\mathcal D}(a)
 &= \frac{\operatorname{PPL}_{\mathcal D}(p)}
          {\operatorname{PPL}_{\mathcal D}(q_a)}\\
 &= \exp\!\bigl(-\widehat\Delta_{\mathcal D}(a)\bigr).
\end{aligned}
\label{eq:ppl-ratio}
\end{equation}
Equal measured likelihood gives $F_{\mathcal D}(a)=1$. During search,
let $U(a)=1$ when execution succeeds and both storage limits hold.
The feedback reward is
\begin{equation}
r_{\mathrm{s}}(a)=
\begin{cases}
10\,F_{\mathcal D_{\mathrm{s}}}(a), & U(a)=1,\\
0, & U(a)=0.
\end{cases}
\label{eq:search-reward}
\end{equation}
Timing is returned separately and does not enter this reward.
In trajectory plots, $E$ abbreviates the measured excess cross-entropy
$\widehat\Delta_{\mathcal D}(a)$, and $F$ denotes the corresponding
PPL ratio when the dataset is clear.

\subsection{Independent Acceptance Tests}
\label{sec:gates}
\label{sec:verification}
Final verification evaluates the frozen weights through the submitted
inference code and official tokenizer. Its contexts are disjoint from
those used for search-time reward. The evaluator controls inputs and
measurements and checks four aspects of the artifact.

The evaluator computes $F_{\mathcal D_{\mathrm{v}}}$ on verification continuations and measures
both package and resident model storage. Retained high-precision copies
count at their actual size. Storage is measured after loading and after
inference to include allocations that persist once the model has run.

Let $\mathcal B$ contain GSM8K, HumanEval, and TriviaQA, with
$A_b(a)$ denoting accuracy on benchmark $b$. Compressed and source models
receive the same questions and generation settings. Mean accuracy and
the acceptance condition are
\begin{equation}
\begin{aligned}
A(a)&=\frac{1}{|\mathcal B|}\sum_{b\in\mathcal B}A_b(a),\\
A(a)&\geq\rho\,A(p),\qquad\rho=0.8,
\end{aligned}
\label{eq:gate-accuracy}
\end{equation}
where $A(p)$ is the BF16 reference fixed before search.
The threshold is set before search and withheld from agents. Held-out answer scores are absent from search-time reward. Separate
questions support the agents' local generation~tests.

\par\vspace{-0.7\baselineskip}\noindent
\begin{minipage}[t]{0.51\linewidth}
\vspace{0pt}\setlength{\parskip}{0.5\baselineskip}
Automated checks and independent review assess compliance with the
construction and inference rules. Tokens and probabilities must come
from the submitted forward pass and packaged weights, with unscaled
logits for likelihood scoring. Review checks model-storage accounting,
external weights, hard-coded outputs, and manipulation of probability
or timing measurements.

A separate test measures elapsed time $L$ between the first and last
of 128 streamed tokens after warmup. We report batch-one decode speed
$v_{\mathrm{dec}}=127/L$ tokens/s: 128 tokens span 127 inter-token intervals.
Prefill and time to first token are excluded. Speed does not enter the
reward, distinguishing compression quality from execution efficiency.

Writing $V(a)=1$ when execution completes, the storage limits and answer
threshold hold, and integrity checks pass, the final normalized score is
\begin{equation}
R(a)=
\begin{cases}
F_{\mathcal D_{\mathrm{v}}}(a), & V(a)=1,\\
0, & V(a)=0.
\end{cases}
\label{eq:reward}
\end{equation}
The final score uses verification data and the full acceptance tests;
search feedback uses $\mathcal D_{\mathrm{s}}$ and $U(a)$. Thus candidate
selection and final acceptance evaluate distinct requirements with
separate data.
\end{minipage}\hfill
\begin{minipage}[t]{0.45\linewidth}
\vspace{0pt}
\captionsetup{hypcap=false}
\centering
\includegraphics[width=\linewidth]{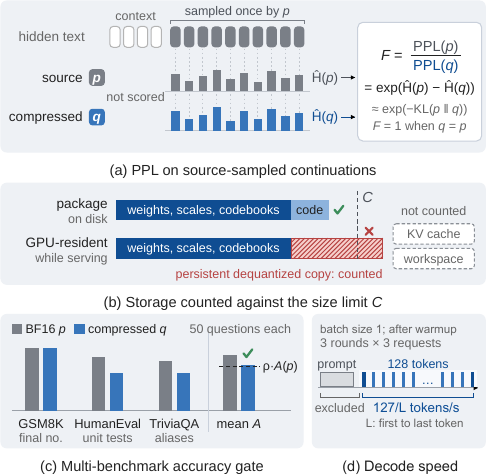}
\captionof{figure}{Verification measurements.
(a) Source $p$ and candidate $q$ score the same source-generated tokens.
(b) Package and resident model storage must each fit $C$, including
persistent reconstructed weights. (c) Mean accuracy over three benchmarks
must retain 80\% of BF16 accuracy, using fifty questions each.
(d) Batch-one decode speed is~$127/L$~tokens/s.}
\label{fig:method}
\end{minipage}
\par

\section{Empirical Study}
\label{sec:evaluation}
We first ask whether agents can deliver useful models under the task's
constraints. We then examine the recorded searches to understand how
agents improve candidates and which failures survive to submission.
\begin{table*}[t]
\centering
\footnotesize
\setlength{\tabcolsep}{3pt}
\renewcommand{\arraystretch}{1.15}
\begin{tabularx}{\textwidth}{@{}l*{3}{>{\raggedleft\arraybackslash}X}@{\hspace{12pt}}*{3}{>{\raggedleft\arraybackslash}X}@{\hspace{12pt}}*{3}{>{\raggedleft\arraybackslash}X}@{}}
\toprule
\multicolumn{10}{@{}l}{\textbf{ (a) Tighter storage limits}}\\[3pt]
 & \multicolumn{3}{c}{\textbf{Nemotron-3.5}} & \multicolumn{3}{c}{\textbf{Qwen3.5}} & \multicolumn{3}{c}{\textbf{Qwen3.8}}\\
 & \multicolumn{3}{c}{120B-A12B\enspace/\enspace\textbf{32 GiB}} & \multicolumn{3}{c}{122B-A10B\enspace/\enspace\textbf{32 GiB}} & \multicolumn{3}{c}{27B\enspace/\enspace\textbf{8 GiB}}\\
\cmidrule(lr){2-4}\cmidrule(lr){5-7}\cmidrule(lr){8-10}
\textbf{Method / agent} & \textbf{PPL}$\downarrow$ & \textbf{Acc.}$\uparrow$ & \textbf{TPS}$\uparrow$ & \textbf{PPL}$\downarrow$ & \textbf{Acc.}$\uparrow$ & \textbf{TPS}$\uparrow$ & \textbf{PPL}$\downarrow$ & \textbf{Acc.}$\uparrow$ & \textbf{TPS}$\uparrow$\\
\rowcolor{RefFill}BF16 & 1.688 & 80.00 & 147.5 & 2.755 & 84.67 & 158.3 & 3.221 & 74.00 & 61.4\\
\rowcolor{RefFill}Released model & \textemdash & \textemdash & \textemdash & 2.900 & 83.33 & 122.9 & 3.764 & 65.33 & 75.5\\
\rowcolor{RefFill}GPTQ & 4.680 & 5.33 & 20.9 & 7.185 & 6.00 & 26.4 & 7.480 & 4.67 & 12.6\\
\addlinespace[3pt]
Sonnet 5.5 & \cellcolor{WinnerFill}\textbf{1.801} & 76.67 & 57.2 & \cellcolor{WinnerFill}\textbf{2.883} & 78.67 & 39.8 & \cellcolor{WinnerFill}\textbf{3.363} & 68.67 & 23.4\\
Opus 5.5 & 1.811 & 85.33 & 48.6 & 3.003 & 76.67 & 73.2 & 3.370 & 66.00 & 26.2\\
GLM 5.3 & 1.871 & 74.67 & 16.6 & \multicolumn{3}{c}{Not passed} & \multicolumn{3}{c}{Not passed}\\
Kimi K3.1 & \multicolumn{3}{c}{Not passed} & 3.035 & 78.00 & 43.1 & \multicolumn{3}{c}{Not passed}\\
MiniMax M3.1 & 2.084 & 75.33 & 11.4 & \multicolumn{3}{c}{Not passed} & \multicolumn{3}{c}{Not passed}\\
DeepSeek V4.1 & \multicolumn{3}{c}{Not passed} & 3.110 & 78.00 & 12.7 & 4.283 & 61.33 & 13.7\\
\addlinespace[5pt]\midrule
\multicolumn{10}{@{}l}{\textbf{ (b) Looser storage limits}}\\[3pt]
 & \multicolumn{3}{c}{\textbf{Nemotron-3.5}} & \multicolumn{3}{c}{\textbf{Qwen3.5}} & \multicolumn{3}{c}{\textbf{Qwen3.8}}\\
 & \multicolumn{3}{c}{120B-A12B\enspace/\enspace\textbf{48 GiB}} & \multicolumn{3}{c}{122B-A10B\enspace/\enspace\textbf{48 GiB}} & \multicolumn{3}{c}{27B\enspace/\enspace\textbf{12 GiB}}\\
\cmidrule(lr){2-4}\cmidrule(lr){5-7}\cmidrule(lr){8-10}
\textbf{Method / agent} & \textbf{PPL}$\downarrow$ & \textbf{Acc.}$\uparrow$ & \textbf{TPS}$\uparrow$ & \textbf{PPL}$\downarrow$ & \textbf{Acc.}$\uparrow$ & \textbf{TPS}$\uparrow$ & \textbf{PPL}$\downarrow$ & \textbf{Acc.}$\uparrow$ & \textbf{TPS}$\uparrow$\\
\rowcolor{RefFill}BF16 & 1.688 & 80.00 & 147.5 & 2.755 & 84.67 & 158.3 & 3.221 & 74.00 & 61.4\\
\rowcolor{RefFill}Released model & 1.764 & 82.67 & 100.5 & 2.808 & 82.00 & 106.9 & 3.334 & 66.00 & 74.6\\
\rowcolor{RefFill}GPTQ & 1.875 & 76.00 & 16.0 & 2.987 & 79.33 & 20.2 & 3.546 & 46.00 & 11.1\\
\addlinespace[3pt]
Sonnet 5.5 & 1.728 & 82.00 & 47.4 & \cellcolor{WinnerFill}\textbf{2.788} & 85.33 & 46.3 & \cellcolor{WinnerFill}\textbf{3.253} & 76.67 & 16.7\\
Opus 5.5 & \cellcolor{WinnerFill}\textbf{1.726} & 78.00 & 68.2 & 2.818 & 82.67 & 52.9 & 3.272 & 78.67 & 31.4\\
GLM 5.3 & 1.778 & 78.67 & 40.3 & 2.814 & 82.67 & 10.5 & 3.383 & 80.67 & 10.4\\
Kimi K3.1 & 1.729 & 82.00 & 13.6 & 2.835 & 84.67 & 7.8 & 3.333 & 74.67 & 8.0\\
MiniMax M3.1 & 1.752 & 81.33 & 9.1 & 2.814 & 84.67 & 27.6 & 3.399 & 76.67 & 10.9\\
DeepSeek V4.1 & 1.749 & 79.33 & 16.6 & 2.803 & 84.67 & 10.8 & 3.399 & 72.00 & 18.4\\
\bottomrule
\end{tabularx}
\caption{Verified models under tighter (a) and looser (b) storage limits.
Perplexity (PPL) is measured on held-out BF16-generated continuations;
accuracy (\%) averages
GSM8K, HumanEval, and TriviaQA. TPS is decode tokens per second at
batch size one, computed as $127/L$ from the 128-token first-to-last
interval $L$. Note that TPS is excluded from the PPL-based reward, and weight unpacking and dequantization can offset the benefits of reduced memory traffic.
Compressed models use one H200 NVL; BF16 uses one for the dense model
and two for the MoE models. Released models use
\href{https://huggingface.co/unsloth/NVIDIA-Nemotron-3.5-Super-VL-120B-A12B-GGUF/tree/main/UD-IQ2_XXS}{Unsloth Dynamic UD-IQ2\_XXS} (Nemotron),
\href{https://huggingface.co/unsloth/Qwen3.5-122B-A10B-GGUF}{Unsloth Dynamic} (Qwen3.5),
and \href{https://huggingface.co/ISTA-DASLab/Qwen3.8-27B-GSQ-RCO-GGUF}{GSQ-RCO} (Qwen3.8). Complete measurements, including rejected artifacts, are in \Cref{tab:artifact-measurements}.}
\label{tab:main-results}
\end{table*}

\subsection{Experimental Setting}
\label{sec:setup}
We study text generation from two mixture-of-experts (MoE) checkpoints,
Nemotron-3.5 Super VL 120B-A12B and Qwen3.5 122B-A10B, and the dense
Qwen3.8 27B checkpoint. Each MoE model has 32 and 48 GiB tasks; the dense
model has 8 and 12 GiB tasks. Pairing limits holds the source model fixed
while changing the need for aggressive compression. Pilot measurements
on Qwen3.8 guide these allowances. Larger pilot limits admit four-bit
GPTQ with group-size or damping refinements; tighter limits make low-bit
representations and precision allocation more consequential. Some pilot
models retain BF16 weights during inference despite small packages,
motivating separate limits on saved and resident model storage
(\Cref{sec:task-calibration}).

Claude Sonnet 5.5, Claude Opus 5.5, GLM 5.3, Kimi K3.1 Pro,
MiniMax M3.1 Flash, and DeepSeek V4.1 Flash 1017 each attempt all six
tasks. Every run uses Claude Code with maximum reasoning effort, one
H200 NVL, vLLM 0.32.0, and an eight-hour budget. Agents receive papers
and FineWeb, UltraChat-200k, and WikiText-103 for calibration or recovery
\citep{penedo2024fineweb,ding2023ultrachat,merity2017wikitext}.
MATH-500 provides separate questions for local answer tests
\citep{lightman2024verify}. There is one run per agent--task pair.

A uniform GPTQ baseline uses groups of 128 weights and 128 calibration
sequences of 2,048 tokens \citep{frantar2023gptq}. It chooses the largest
uniform bit width that fits the package limit and adjusts embedding and
output-head precision to the remaining space. We also evaluate released
quantizations of the same checkpoints: Unsloth Dynamic GGUF for Nemotron
and Qwen3.5, and GSQ-RCO for Qwen3.8
\citep{dadgarnia2026gsq,helcig2026compression}.

All compressed models run on one GPU. Agent artifacts and GPTQ use
vLLM \citep{kwon2023vllm} with packed weights; published weights run unchanged in their
native packed GGUF representation using llama.cpp.
BF16 uses two GPUs for the MoE models and one for the dense model.
All comparisons share the measurement protocol; TPS also reflects the
submitted kernel implementations and dequantization overhead.

Search-time and verification likelihood each use 256 contexts of 512
tokens from the Dolma 3 Dolmino mixture \citep{olmo2025olmo3}.
The sets use disjoint documents with matched source proportions.
For each context, BF16 samples 1,536 tokens at temperature one without
probability truncation; both models score those continuations.
We report final PPL in tables and search-time feedback in trajectory
plots. Cross-model comparisons use PPL increase over BF16.

Answer accuracy is the mean over fifty held-out questions each from
GSM8K, HumanEval, and TriviaQA
\citep{cobbe2021gsm8k,chen2021humaneval,joshi2017triviaqa}, with
identical questions and greedy decoding for source and compressed models.
Acceptance uses BF16 references fixed before search. The main table
also reports fresh BF16 measurements; these remeasurements supply the
77.33\% versus 79.56\% aggregate comparison below.

Verification runs on an idle GPU and counts model allocations after
loading and inference. Decode uses batch size one, a 2,048-token prompt,
and 128 output tokens. After warmup, $L$ is the median of three round
means over three requests each, timed between first and last token.
Reported decode speed is $127/L$ tokens/s. The fixed BF16 references, generation budgets, measurement protocol,
per-artifact results, and additional experiments appear in
\Cref{sec:trajectory-cohorts,sec:intermediate-verification,sec:additional-studies}.

\subsection{Delivered Models}
\label{sec:main-results}
\label{sec:size-results}
Twenty-nine of the thirty-six submissions pass final verification
(\Cref{tab:main-results}). All eighteen at the looser limits pass.
At the tighter limits, six fail the answer-accuracy requirement and one
times out during answer generation. Selecting the lowest-PPL agent
artifact for each task gives an average 3.4\% PPL increase over BF16.
Those six artifacts average 77.33\% downstream accuracy, compared with
79.56\% for the remeasured BF16 models.

The storage allowance influences every agent's result, but does not
fully explain quality. Each agent's larger-allowance artifact has lower
PPL than its tighter counterpart. Yet Sonnet's 8 GiB dense model also
has lower PPL than three agents' 12 GiB models. A better representation
can therefore outweigh additional storage in these comparisons. Uniform
GPTQ passes the answer threshold on the looser MoE tasks and fails on
the remaining settings, indicating that a fixed recipe covers only part of the task~suite.

Released quantizations provide a stronger comparison. The best agent
artifact by PPL improves on each of the five measured released-model
references. Answer accuracy distinguishes designs further: the tighter
Qwen3.5 release has higher accuracy than the lowest-PPL agent model,
whereas Sonnet improves both PPL and accuracy over the released dense
references. PPL and answer accuracy thus affect
which artifact would be preferred.

\Cref{fig:tradeoff-teaser} places the results alongside encoded size and
API cost. Sonnet has the best average PPL rank, followed by Opus;
the largest API expenditure does not correspond to the best rank.
The remaining analysis follows the concrete decisions behind these
outcomes, rather than using spending or activity alone to explain them.

\section{How Agents Improve Compression}
\label{sec:analysis}
\label{sec:mechanisms}
Agents can change compression algorithms, storage allocation across
model components, or candidate evaluation. Recorded trajectories
connect these choices to model measurements and subsequent actions.
Within-task comparisons follow the recorded revisions; a paired
ablation isolates a quantizer decision after search.

\subsection{Representation and Storage Tradeoffs}
\label{sec:numerical-design}
\label{sec:compression-search}
\begin{figure}[t]
\centering
\includegraphics[width=\textwidth]{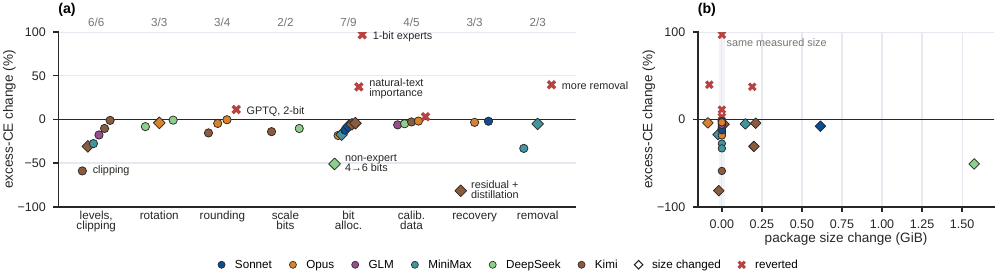}
\caption{Revisions within recorded searches.
Thirty-five pairs from 21 runs compare candidate PPL through reward
feedback. Excess cross-entropy is $E=\ln(\operatorname{PPL}/\operatorname{PPL}_{\mathrm{BF16}})$;
the vertical axis is $100(E_{\mathrm{after}}/E_{\mathrm{before}}-1)$.
(a) Changes by design factor, with lower-PPL/total counts.
(b) The same pairs by package-size change; the band denotes changes
within 0.01 GiB, and crosses mark candidates subsequently~reverted.}
\label{fig:representation}
\end{figure}

The first question is where an agent finds improvement once a model
fits the storage limit. It can allocate more bytes to a sensitive
component only by saving them elsewhere, or improve the approximation
represented by existing codes. In 24 recorded
revisions whose package sizes differ by at most 0.01 GiB, 21 lower PPL
measured by reward feedback. These pairs span all six agents and all six tasks
 (\Cref{fig:representation}). These choices improve the
quantizer or redistribute precision within a feasible model.

On tighter Qwen3.5, Kimi changes clipping and GPTQ damping while keeping
package size unchanged, reducing feedback PPL from 3.738 to 3.132.
Across the trajectories, level and clipping revisions improve PPL for
three agents on five tasks. Bit width fixes the code count; clipping and rounding determine how
those codes approximate weights and activations
\citep{frantar2023gptq,shao2024omniquant}. The paired dense-model
ablation in \Cref{sec:paired-ablations} keeps precision allocation,
calibration, and inference implementation fixed while replacing
GLM's MSE-selected clipping ranges with min--max ranges.
The selected ranges give lower verification PPL and higher answer
accuracy. Both variants occupy the same saved and resident model
storage. All three answer benchmarks favor the selected ranges. This connects the agent's quantizer choice to the quality of the complete model after search.

Sensitivity guides mixed-precision bit allocation
\citep{dong2020hawqv2}, but agents also allocate storage to metadata.
Kimi reduces dense-model group-scale storage to fit additional four-bit
weights. DeepSeek raises
non-expert precision on Nemotron, while Opus gives later Qwen3.5 layers
more expert precision. Weights and metadata share the package budget. Kimi reduces the storage used for scales to make room for higher weight precision; DeepSeek and Opus change where that precision is assigned. These examples show why the allocation includes quantizer overhead as well as the nominal bit width of the weights.

An agent can also remove parameters to afford higher precision for
those that remain. GLM and MiniMax test additional expert or MLP-width
removal; their recorded comparisons favor the less-pruned alternatives.
MiniMax nevertheless retains a large dense-model width reduction after
rejecting an early two-bit implementation. Its subsequent generation
tests expose a substantial loss. MiniMax's decision to retain pruning
links its precision choice to parameter removal. Subsequent numerical
refinements operate on a model with fewer parameters.

\FloatBarrier
\subsection{Local Metrics and Candidate Selection}
\label{sec:validation-design}
\label{sec:answer-accuracy}
\begin{wrapfigure}{r}{0.45\linewidth}
\centering
\includegraphics[width=\linewidth]{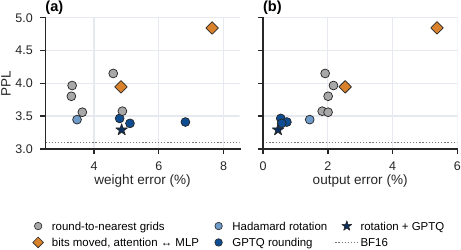}
\caption{Local error and model likelihood.
Twelve Qwen3.8 variants use approximately 3.14 encoded bits per quantized
decoder weight, with BF16 embeddings and output head. PPL is measured
using BF16 reconstructions on the source continuations used for reward
feedback. (a) Weight reconstruction MSE. (b) Linear-output MSE on BF16
FineWeb calibration activations. Each error is normalized by its BF16
squared norm; the dotted line denotes BF16 PPL.}
\label{fig:quantizer-design}
\end{wrapfigure}
The second question is how an agent chooses which design to retain.
Local experiments guide which candidate quantizers and precision
allocations an agent builds. We examine what happens when reward
feedback disagrees with that local ranking: whether the agent
reverts the candidate, revises its numerical choices, or changes
the validation used in subsequent experiments. Across 79 matched comparisons from 26 runs, reward feedback reverses
17 prior local rankings across all six tasks
and all agents except MiniMax. Eight reversals lead to explicit rollback: five by Kimi and
three by DeepSeek. Across the eligible comparisons, agents also refine
calibration and quantization or reallocate precision
(\Cref{sec:agent-response-analysis}). Model-level measurements thus
influence both the candidate an agent retains and the numerical choices
it revisits.

A controlled diagnostic on dense Qwen3.8 varies levels, clipping,
rotation, rounding, and attention--MLP bit allocation at similar encoded
size (\Cref{fig:quantizer-design}). Forward passes use BF16 reconstructions,
separating numerical representation from packed-kernel execution.
A k-means codebook improves weight reconstruction MSE over clipped
uniform levels but increases PPL. GPTQ rounding has the opposite
pattern: larger weight MSE and lower PPL. Calibration-activation output error follows PPL more closely in this
comparison. Weight MSE treats reconstruction errors equally across input
directions. Linear-output MSE weights those errors by the calibration
activations. A small reconstruction error in a direction with large
activation magnitudes can affect the layer's output more than a larger
error in a weakly activated direction. Complete-model PPL measures the
resulting behavior after all quantized layers are composed. These criteria can therefore favor different
revisions at similar encoded size.

\suppressfloats[t]
\begin{figure*}[t]
\centering
\includegraphics[width=\textwidth]{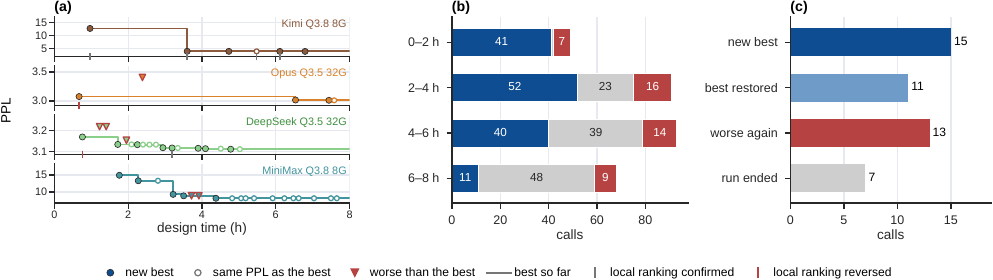}
\caption{Feedback during candidate selection.
(a) Best-so-far PPL for four searches; ticks indicate whether complete-model
feedback confirms or reverses a local ranking. (b) Outcomes of 301 reward
evaluations across 36 runs, grouped by search time. Rewards within 0.002
of the best are treated as the same PPL. (c) The next outcome after 46
worse candidates. Reward equals ten times the source-to-compressed PPL
ratio; all plotted evaluations occur during search.}
\label{fig:validation}
\end{figure*}

On looser Qwen3.5, Opus initially favors more precision for early
layers. It revises local validation to score model-generated
continuations, then favors later layers, an allocation supported by
reward feedback. On the tighter task, it spends 1.7 hours exploring
a one-bit expert tier. Local likelihood favors the new candidate, but
feedback favors the previous allocation; Opus restores and refines that
allocation. Conflicting measurements lead it to revise local validation
in one search and revert the candidate in the other.

Later evaluations increasingly match a run's best PPL
(\Cref{fig:validation}). Some confirm that a candidate is worth keeping;
others follow successful numerical revisions. Kimi improves its dense
model after adding a low-rank residual and distillation. MiniMax instead
repeatedly evaluates a pruned candidate after its last improvement.
Reward feedback reports a tested candidate's performance; the agent
then decides whether to refine its quantizer, change precision
allocation, or reconsider the representation.

\subsection{Validating Generation and Execution}
\label{sec:capacity-precision}
\label{sec:agentic-design}
A prior pruning study reports mismatches between perplexity and
downstream performance \citep{khanal2024compression}. For agents, detecting a generation
regression and finding a feasible repair are distinct challenges. PPL scores fixed source-generated histories, while generation feeds
the candidate's own choices back into the model. Repetition, early
termination, and changed problem solving therefore require direct
checks. Intermediate verification tracks these changes between
versions; trajectories show whether agents detect the loss and
find a repair they can retain.

\begin{figure*}[t]
\centering
\includegraphics[width=\textwidth]{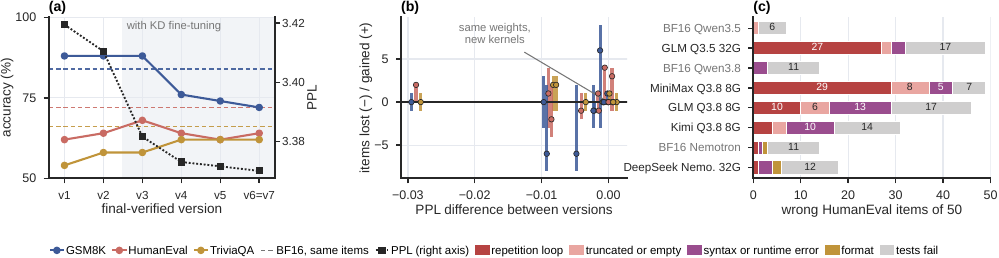}
\caption{Capability changes across candidate models.
(a) Verification PPL and answer accuracies for Opus on Qwen3.8 / 8 GiB;
shading marks distillation and dashed lines mark BF16. (b) Answers gained
and lost across twelve successive-version pairs from four runs, plotted
against PPL change. (c) HumanEval failure types for selected rejected
artifacts and their BF16 references. Each benchmark has fifty questions.}
\label{fig:capacity-tradeoff}
\end{figure*}

We verify retained intermediate models after search using the same
held-out inputs as the final artifacts. In Opus's tighter dense-model
run, allocation and distillation improve PPL while GSM8K accuracy falls
from 88\% to 72\% across trained versions. HumanEval and TriviaQA improve
over the same versions. Twelve comparisons from four runs show both
newly correct and newly incorrect answers after small PPL changes
(\Cref{fig:capacity-tradeoff}). A single scalar can conceal changes in
which capabilities survive.

Generation failures also differ in kind. Pruned GLM and MiniMax models
produce repeated or truncated HumanEval completions, while most BF16
errors are executable programs that fail tests. This matters for
acceptance: an artifact may preserve scored-token likelihood yet become
unreliable at completing an~answer.

\begin{figure}[t]
\centering
\includegraphics[width=\textwidth]{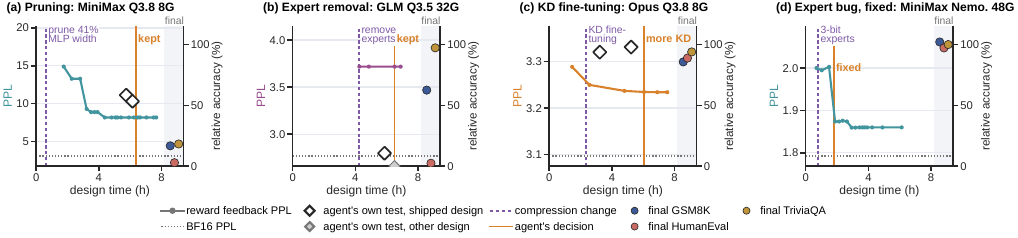}
\caption{Testing and acting on failures.
MiniMax's dense-model run (a) and GLM's Qwen3.5 run (b) retain pruning;
Opus continues dense-model distillation (c); MiniMax repairs expert
indexing on Nemotron (d). Curves show feedback PPL. Diamonds show the
agents' generation tests relative to an unpruned three-bit reference (a),
an eight-bit reference (b), or BF16 (c). Final markers show held-out
benchmark accuracy relative to BF16.}
\label{fig:gate-accuracy}
\end{figure}

MiniMax's local reasoning tests show a large loss after MLP-width
pruning, relative to its unpruned three-bit reference. GLM's expert-pruned
Qwen3.5 candidate similarly underperforms an eight-bit reference on
MATH-500. Both agents test alternatives and retain pruning without
finding a feasible design that resolves the loss. Opus tests generation
during distillation but makes later PPL-oriented changes without
repeating the same generation test on its last versions. Those local
MATH-500 measurements and the post-search GSM8K loss are separate tests.
The trajectories expose two distinct requirements: finding a candidate
that fixes a known problem and keeping tests current with the artifact
ultimately submitted.

On Nemotron, MiniMax compares token probabilities under batched and
per-expert inference. The disagreement identifies an expert-indexing
bug. Repair reduces the feedback PPL increase over BF16 from about
11\% to 4.5\%, and the final model passes verification. Here, the agent repairs
execution while keeping the representation fixed. In the pruning cases,
generation tests reveal regressions, but agents retain the affected
designs. Detecting a failure and finding a feasible change to the
compression design or implementation that resolves it are distinct
challenges.

\subsection{Two Complete Compression Strategies}
\label{sec:case-study}
\begin{figure*}[t]
\centering
\includegraphics[width=\textwidth]{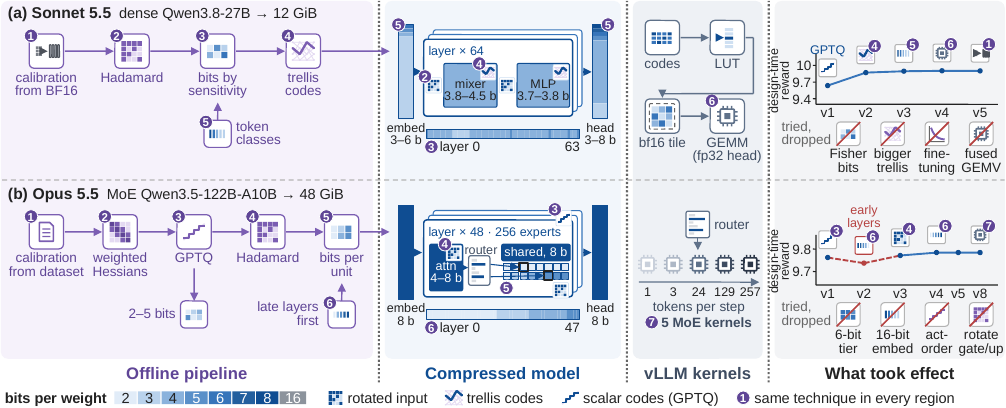}
\caption{From numerical design to execution.
Sonnet compresses dense Qwen3.8 to 12 GiB using rotation, trellis codes,
and sensitivity-based allocation (a). Opus compresses MoE Qwen3.5 to
48 GiB using rotated GPTQ and expert-specific precision (b). Rows connect
the offline pipeline, component precision, inference operations, and
candidate versions. Curves show search-time reward; boxes list explored
alternatives that were discarded.}
\label{fig:case-pipeline}
\label{fig:case-progress}
\end{figure*}

The dense and MoE case studies in \Cref{fig:case-pipeline} bring these
decisions together. Both retain active parameters at the looser limits,
but their numerical methods lead to different allocations and execution
paths.

\noindent\textbf{\textbullet\ Sonnet: trellis codes and bit allocation.}
Sonnet replaces scalar GPTQ with Hadamard-rotated trellis codes, error
compensation, and Viterbi search \citep{frantar2023gptq,tseng2024qtip}. Matrix experiments
help choose the representation. KL ablations supply a fitted loss model
for bit allocation; token frequency guides embedding and output-head
precision. Representation and allocation changes produce the largest
reward gain, followed by calibration refinements. Sonnet rejects
Fisher-weighted allocation when its measured KL worsens and keeps a
smaller trellis when a larger one offers little reconstruction gain at
greater search~cost. Its final artifact has PPL 3.253 and 76.67\% mean
answer accuracy, versus BF16's 3.221 and 74.00\%.

\noindent\textbf{\textbullet\ Opus: expert-specific precision.}
Opus uses rotated GPTQ with different bit widths across experts.
Revising validation shifts the favored allocation toward later layers,
while natural text remains the calibration source. Its search rejects
additional precision tiers and alternative rotations. The final artifact
has PPL 2.818 and 82.67\% accuracy, versus BF16's 2.755 and 84.67\%.

Intermediate verification separates likelihood from answer preservation.
Sonnet's versions v3, v4, and v5 attain PPL 3.2572, 3.2531, and 3.2525,
but answer 111, 104, and 115 of 150 questions correctly. Opus's Qwen3.5
version v4 has PPL 2.8172 and 120 correct answers; v5--v8 have PPL 2.8178
and 124 correct answers. These tests use final verification inputs;
search-time feedback guides candidate selection. Nearly unchanged
likelihood thus accompanies different generated answers.

\label{sec:inference-results}
\label{sec:execution}
Sonnet reconstructs weight tiles from its stored trellis codes as they are needed for matrix computation. Other designs can first reconstruct a whole weight
matrix in temporary workspace and then use dense operations. Both
approaches can satisfy the resident-storage limit because the
reconstructed values are temporary. They expose different amounts
of unpacking work and memory traffic to inference, which contribute
to the measured execution cost. In the dense looser task,
Sonnet improves PPL and accuracy over the released GSQ-RCO artifact,
but decodes at 16.7 tokens/s versus the release's 74.6 tokens/s
at batch size one
(\Cref{tab:artifact-measurements}). This illustrates why the measured
storage--quality improvement does not by itself establish a speedup,
and why representation and kernels must be considered together
\citep{lin2024awq,lin2025qserve}.

\keepspace{6\baselineskip}   
\section{Related Work}
\label{sec:related-work}
\subsection{Agentic Benchmarks}
AlphaEvolve optimizes algorithms and LLM-training infrastructure with
coding agents \citep{novikov2025alphaevolve}. The Darwin G\"odel Machine
modifies agent code and evaluates the resulting coding agents
\citep{zhang2026darwin}. We study a complementary efficiency objective:
compressed model weights and their execution under a resource budget.
The terminal task connects numerical choices and validation decisions
to the quality of the executable~model.

Terminal-Bench evaluates agents on verified command-line tasks
\citep{merrill2026terminalbench}; Harbor provides a container-based
framework for executing tasks and evaluating agents \citep{harbor2026}.
SWE-bench studies repository repair \citep{jimenez2024swebench}, and
AgentBench evaluates interactive environments \citep{liu2024agentbench}.
MLAgentBench, MLE-bench, and RE-Bench examine ML experimentation,
competition solutions, and research engineering
\citep{huang2024mlagentbench,chan2025mlebench,wijk2025rebench}.
KernelBench tests correctness and runtime for specified tensor computations
\citep{ouyang2025kernelbench}; ServeLearnBench studies learning from
serving feedback \citep{zheng2026servelearnbench}. Compression adds the
choice of numerical approximation: the agent must select, implement,
and validate a model whose generated answers remain useful.

\subsection{LLM Compression}
GPTQ, AWQ, SmoothQuant, and OmniQuant use calibration to choose
rounding, scaling, or clipping
\citep{frantar2023gptq,lin2024awq,xiao2023smoothquant,shao2024omniquant}.
QuIP, QuaRot, and SpinQuant improve quantization through incoherence
processing or rotations
\citep{chee2023quip,ashkboos2024quarot,liu2025spinquant}; QuIP\#, AQLM,
and QTIP develop lattice, additive, or trellis representations
\citep{tseng2024quipsharp,egiazarian2024aqlm,tseng2024qtip}.
HAWQ-V2 uses sensitivity for mixed precision \citep{dong2020hawqv2}, while
GSQ and RCO optimize quantization and budget allocation
\citep{dadgarnia2026gsq,helcig2026compression}.
Pruning removes weights or dimensions; low-rank methods reduce matrix rank
\citep{frantar2023sparsegpt,sun2024wanda,ashkboos2024slicegpt,wang2025svdllm}.
Recovery can use quantization-aware fine-tuning
\citep{malinovskii2024pvtuning} or distillation
\citep{hinton2015distilling,muralidharan2024compact}.
Serving systems optimize memory management and attention execution
\citep{kwon2023vllm,ye2025flashinfer}; TinyChat and QServe couple quantized
formats with matching kernels \citep{lin2024awq,lin2025qserve}.

These methods supply components; we study agents that select, combine,
implement, and validate a complete compressed model.

\section{Conclusion}
\label{sec:conclusion}
LLM compression provides a concrete task for studying agent-driven model
improvement. Across 36 runs, 29 artifacts pass independent verification.
Recorded searches show quality gains at nearly fixed storage, local
rankings overturned by complete-model feedback, and generation or
implementation failures missed by likelihood alone. Effective compression
requires numerical design, experimental judgment, and validation of the
final executable model. Extending this step toward RSI requires testing
whether a compressed model retains the ability to carry out subsequent
compression~tasks.

\section*{Acknowledgements}
This work was partially supported by Google Research Award, Google ML \& System Junior Faculty Award, Amazon Research Award, Fireworks AI, Intel, Li Auto, Moffett AI, Daytona, and CMU CyLab Seed funding. This material is also based upon work supported by the National Science Foundation under Grant Nos. CCF-2504353 and CCF-2247014, and by IARPA. Any opinions, findings, conclusions or recommendations expressed are those of the authors and do not necessarily reflect the views of the National Science Foundation.

\clearpage
\bibliographystyle{assets/plainnat}
\bibliography{references}

\section*{AI Usage Disclosure}
Claude Code and OpenAI Codex assisted with manuscript rewriting, literature and citation
checks, LaTeX preparation, and PDF compilation. The compression artifacts
and numerical measurements come from the recorded experimental runs
described in the paper.

\clearpage
\beginappendix
\setcounter{figure}{0}
\setcounter{table}{0}
\renewcommand{\thefigure}{A\arabic{figure}}
\renewcommand{\thetable}{A\arabic{table}}
\renewcommand{\theHfigure}{appendix.\arabic{figure}}
\renewcommand{\theHtable}{appendix.\arabic{table}}
\section{Evaluation Protocol}
\label[appendix]{sec:trajectory-cohorts}
Each likelihood set contains 256 contexts of 512 tokens from Dolma 3
Dolmino, stratified by source family. BF16 samples 1,536 continuation
tokens at temperature one without probability truncation; both models score the
393,216 tokens. Reward feedback and verification use disjoint documents
with matched source proportions. \Cref{tab:bf16-reference} gives the
references fixed before agentic design.

\begin{table}[htbp]
\centering\small
\begin{tabular}{@{}lrrr@{}}
\toprule
Source & \shortstack{Reward\\feedback PPL} & \shortstack{Verification\\PPL} & $A(p)$ (\%)\\
\midrule
Nemotron-3.5 & 1.792648 & 1.688011 & 81.33\\
Qwen3.5 & 2.768058 & 2.755500 & 85.33\\
Qwen3.8 & 3.097426 & 3.220743 & 74.00\\
\bottomrule
\end{tabular}
\caption{BF16 references fixed before agentic design.
PPL is perplexity on disjoint source-generated continuations for reward feedback and verification. $A(p)$ is mean accuracy (\%) on GSM8K, HumanEval,
and TriviaQA, used to set the accuracy gate.}
\label{tab:bf16-reference}
\end{table}

Source and compressed models use the same questions, greedy decoding,
and settings in \Cref{tab:gate-details}. Acceptance requires
$A(a)\geq0.8A(p)$, with the fixed reference in
\Cref{tab:bf16-reference}.

\begin{table}[htbp]
\centering\small
\begin{tabular}{@{}lrrl@{}}
\toprule
Dataset & Shots & Tokens & Judgment\\
\midrule
GSM8K & 4 & 320 & Final number\\
HumanEval & 0 & 384 & Unit tests\\
TriviaQA & 5 & 24 & Normalized aliases\\
\bottomrule
\end{tabular}
\caption{Multi-benchmark accuracy protocol.
Each benchmark uses fifty held-out questions and greedy decoding.
Shots gives in-context examples; Tokens gives maximum answer length.
The three benchmark accuracies are averaged equally.}
\label{tab:gate-details}
\end{table}

Package size includes weights, inference code, and tokenizer assets.
Resident storage counts persistent weights, quantization metadata, and
reconstructed copies after loading and inference. Shared allocations
count once; KV caches, recurrent states, and temporary workspace are
excluded.

Batch-one decode uses 2,048 input and 128 output tokens after warmup.
TPS is $127/L$, where $L$ is the median of three round means, each over
three first-to-last-token intervals. BF16 speeds are 147.5, 158.3, and
61.4 tokens/s for Nemotron, Qwen3.5, and Qwen3.8, respectively.
Final verification has a 3,600-second timeout.

Candidate-ranking pairs match local rankings with subsequent reward
feedback for the same configurations. Package bits per original
parameter equal eight times package bytes divided by parameter count;
component precision uses resident bytes and component parameter count.
API cost uses recorded input, output, and cached tokens at October 26,
2026 OpenRouter rates. Average ranks weight all six tasks equally.

\section{Search Activity and Stopping}
\begin{figure*}[t]
\centering
\includegraphics[width=0.91\textwidth]{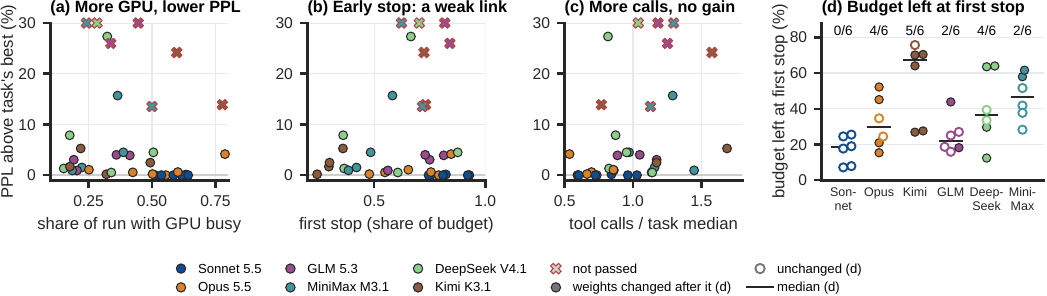}
\caption{Computation and stopping during agentic design.
(a--c) Final-verification perplexity (PPL) increase over the best submission on the same
task versus GPU-busy share of the run, first completion declaration as a
fraction of the design budget, and tool calls divided by the task median.
Values above 30\% lie at the upper boundary. (d) Budget remaining at the
first completion declaration for each of thirty-six runs. Filled points
mark subsequent weight changes; counts give changed/total runs, and bars
show medians. Agents often revise weights after declaring completion.}
\label{fig:budget-outcome}
\end{figure*}

\label[appendix]{sec:agentic-design-time}
We examine how agents use the design budget and when they stop revising
weights (\Cref{fig:budget-outcome}). Agents were instructed to carry out
agentic design end to end until the time limit. Runs proceeded without
human intervention or follow-up prompts.

Seventeen runs change weights after the agent first declares the task
complete: five Kimi runs, four each for Opus and DeepSeek, two each for
GLM and MiniMax, and none for Sonnet. Kimi has the earliest median first
completion declaration and usually continues revising its weights. Median GPU-busy shares range from 30\% to 60\% across
agents. These records distinguish a declaration of completion from the
last effective model change.

\section{Final and Intermediate Model Measurements}
\label[appendix]{sec:intermediate-verification}
\Cref{tab:artifact-measurements} contains the full final measurements,
including models rejected by the answer threshold and decode TPS for both
compression baselines.
\begin{table}[tp]
\centering\footnotesize
\setlength{\tabcolsep}{3pt}
\renewcommand{\arraystretch}{1.0}
\begin{tabular*}{\columnwidth}{@{\extracolsep{\fill}}lrrrrrc@{}}
\toprule
\textbf{Method} & \textbf{PPL}$\downarrow$ & \textbf{Acc.}$\uparrow$ & \textbf{Package} & \textbf{Resident} & \textbf{TPS}$\uparrow$ & \textbf{Gate}\\
\midrule
\rowcolor{RefFill}\multicolumn{7}{@{}l@{}}{\strut\textbf{Nemotron-3.5 120B-A12B, 32 GiB}}\\
GPTQ & 4.680 & 5.33 & 31.959 & 31.980 & 20.9 & Fail\\
Sonnet & \textbf{1.801} & 76.67 & 31.665 & 31.649 & 57.2 & Pass\\
Opus & 1.811 & 85.33 & 31.943 & 31.912 & 48.6 & Pass\\
GLM & 1.871 & 74.67 & 31.873 & 31.841 & 16.6 & Pass\\
Kimi & 2.051 & 58.67 & 31.999 & 31.975 & 12.5 & Fail\\
MiniMax & 2.084 & 75.33 & 30.930 & 30.905 & 11.4 & Pass\\
DeepSeek & 2.369 & 63.33 & 31.844 & 31.814 & 13.8 & Fail\\
\midrule
\rowcolor{RefFill}\multicolumn{7}{@{}l@{}}{\strut\textbf{Nemotron-3.5 120B-A12B, 48 GiB (release: UD-IQ2\_XXS)}}\\
GPTQ & 1.875 & 76.00 & 45.981 & 45.995 & 16.0 & Pass\\
Unsloth & 1.764 & 82.67 & 47.625 & 44.685 & 100.5 & Pass\\
Sonnet & 1.728 & 82.00 & 47.850 & 47.834 & 47.4 & Pass\\
Opus & \textbf{1.726} & 78.00 & 47.904 & 47.854 & 68.2 & Pass\\
GLM & 1.778 & 78.67 & 47.232 & 47.207 & 40.3 & Pass\\
Kimi & 1.729 & 82.00 & 47.958 & 47.915 & 13.6 & Pass\\
MiniMax & 1.752 & 81.33 & 47.686 & 47.670 & 9.1 & Pass\\
DeepSeek & 1.749 & 79.33 & 47.689 & 47.673 & 16.6 & Pass\\
\midrule
\rowcolor{RefFill}\multicolumn{7}{@{}l@{}}{\strut\textbf{Qwen3.5 122B-A10B, 32 GiB (release: UD-IQ1\_M)}}\\
GPTQ & 7.185 & 6.00 & 31.604 & 31.648 & 26.4 & Fail\\
Unsloth & 2.900 & 83.33 & 31.883 & 31.861 & 122.9 & Pass\\
Sonnet & \textbf{2.883} & 78.67 & 31.746 & 31.765 & 39.8 & Pass\\
Opus & 3.003 & 76.67 & 31.728 & 31.750 & 73.2 & Pass\\
GLM & 3.632 & 44.67 & 31.715 & 31.734 & 18.3 & Fail\\
Kimi & 3.035 & 78.00 & 31.918 & 31.937 & 43.1 & Pass\\
MiniMax & 3.273 & \textemdash & 31.510 & \textemdash & \textemdash & Timeout\\
DeepSeek & 3.110 & 78.00 & 31.840 & 31.832 & 12.7 & Pass\\
\midrule
\rowcolor{RefFill}\multicolumn{7}{@{}l@{}}{\strut\textbf{Qwen3.5 122B-A10B, 48 GiB (release: UD-IQ3\_S)}}\\
GPTQ & 2.987 & 79.33 & 47.133 & 47.173 & 20.2 & Pass\\
Unsloth & 2.808 & 82.00 & 43.372 & 43.349 & 106.9 & Pass\\
Sonnet & \textbf{2.788} & 85.33 & 47.669 & 47.689 & 46.3 & Pass\\
Opus & 2.818 & 82.67 & 47.847 & 47.866 & 52.9 & Pass\\
GLM & 2.814 & 82.67 & 47.688 & 47.707 & 10.5 & Pass\\
Kimi & 2.835 & 84.67 & 47.976 & 47.995 & 7.8 & Pass\\
MiniMax & 2.814 & 84.67 & 47.658 & 47.770 & 27.6 & Pass\\
DeepSeek & 2.803 & 84.67 & 47.686 & 47.706 & 10.8 & Pass\\
\midrule
\rowcolor{RefFill}\multicolumn{7}{@{}l@{}}{\strut\textbf{Qwen3.8 27B, 8 GiB (release: IQ2\_XS)}}\\
GPTQ & 7.480 & 4.67 & 7.354 & 7.515 & 12.6 & Fail\\
GSQ-RCO & 3.764 & 65.33 & 7.856 & 7.834 & 75.5 & Pass\\
Sonnet & \textbf{3.363} & 68.67 & 7.957 & 7.950 & 23.4 & Pass\\
Opus & 3.370 & 66.00 & 7.977 & 7.995 & 26.2 & Pass\\
GLM & 6.061 & 23.33 & 7.947 & 7.953 & 13.7 & Fail\\
Kimi & 4.176 & 54.00 & 7.973 & 7.992 & 7.9 & Fail\\
MiniMax & 8.163 & 9.33 & 7.973 & 7.840 & 15.1 & Fail\\
DeepSeek & 4.283 & 61.33 & 7.934 & 7.950 & 13.7 & Pass\\
\midrule
\rowcolor{RefFill}\multicolumn{7}{@{}l@{}}{\strut\textbf{Qwen3.8 27B, 12 GiB (release: IQ3\_S)}}\\
GPTQ & 3.546 & 46.00 & 11.401 & 11.531 & 11.1 & Fail\\
GSQ-RCO & 3.334 & 66.00 & 10.975 & 10.953 & 74.6 & Pass\\
Sonnet & \textbf{3.253} & 76.67 & 11.956 & 11.977 & 16.7 & Pass\\
Opus & 3.272 & 78.67 & 11.980 & 11.970 & 31.4 & Pass\\
GLM & 3.383 & 80.67 & 11.954 & 11.973 & 10.4 & Pass\\
Kimi & 3.333 & 74.67 & 11.967 & 11.986 & 8.0 & Pass\\
MiniMax & 3.399 & 76.67 & 11.779 & 11.798 & 10.9 & Pass\\
DeepSeek & 3.399 & 72.00 & 11.948 & 11.967 & 18.4 & Pass\\
\bottomrule
\end{tabular*}
\caption{Complete final-verification measurements by source model and
storage limit. PPL scores held-out BF16-generated continuations;
accuracy (\%) averages GSM8K, HumanEval, and TriviaQA. Package and
resident model storage are in GiB. TPS is batch-one decode speed,
$127/L$ for a 128-token first-to-last interval $L$. The accuracy gate
requires at least 80\% of the fixed BF16~reference.}
\label{tab:artifact-measurements}
\end{table}

Intermediate models are evaluated after search on the same verification
inputs as the final submissions. The main-paper trajectory curves report
reward feedback gathered during agentic design.

On Qwen3.8 at 8 GiB, Opus's reallocation lowers verification PPL
increase from 6.17\% to 5.89\%. Distillation lowers it to 5.00\%,
then further allocation and training reach 4.64\%. GSM8K accuracy
falls from 88\% to 72\% across trained versions, while the other
benchmark changes are smaller.

\section{Storage Limits and Reference Quantizations}
\label[appendix]{sec:task-calibration}
Pilot compression runs on Qwen3.8 27B inform the task allowances and
storage accounting. Pilot PPL is measured on 256 held-out
C4~\citep{raffel2020t5} contexts of 2,048 tokens, with BF16 PPL 9.9227.
The main study instead evaluates source-generated continuations.

\Cref{tab:calib-limits} compares compression choices across four storage
allowances. Pilot timing follows the first-to-last-token protocol in
\Cref{sec:trajectory-cohorts} with an earlier vLLM release.

\begin{table}[t]
{
\centering\footnotesize
\setlength{\tabcolsep}{2pt}
\begin{tabularx}{\columnwidth}{@{}Xrrr@{}}
\toprule
Submission & \shortstack{Package\\(GiB)} & \shortstack{PPL above\\BF16} & \shortstack{TPS$\uparrow$\\(bsz=1)}\\
\midrule
\multicolumn{4}{l}{\textit{32 GiB}}\\
GLM: 4-bit GPTQ/Marlin & 17.34 & +3.09\% & 106.1\\
GLM: 4-bit GPTQ (no example) & 17.34 & +3.22\% & 106.0\\
Sol: optimized GPTQ & 16.43 & +3.38\% & 111.9\\
GLM: 5-bit, group 64 & 20.52 & +0.60\% & 94.8\\
\midrule
\multicolumn{4}{l}{\textit{16 GiB}}\\
4-bit, group 128, damping 0.01 & 15.00 & +3.19\% & 109.0\\
4-bit, group 64 & 15.35 & +2.58\% & 108.6\\
4-bit, group 32 (over limit) & 16.06 & +1.98\% & 107.6\\
4-bit, group 128, damping 0.1 & 15.00 & +2.79\% & 109.3\\
GLM A: group 64 for MLP & 15.83 & +3.89\% & 109.9\\
GLM B: 4-bit embedding/head & 14.99 & +3.10\% & 109.1\\
\midrule
\multicolumn{4}{l}{\textit{12 GiB}}\\
3-bit, group 128 & 11.18 & +13.32\% & 110.6\\
Mixed 3/4-bit, group 128 & 11.98 & +10.53\% & 110.8\\
GLM A: 3/4-bit (load failure) & 11.83 & -- & --\\
GLM B: Marlin repacking & 11.97 & +9.43\% & 111.1\\
\midrule
\multicolumn{4}{l}{\textit{8 GiB}}\\
GLM: Hadamard, 2/3/4-bit & 7.96 & +1739\% & 61.9\\
Qwen: 2/3/4-bit, clipping & 7.94 & +48.03\% & 62.0\\
MiMo: 2/3/4-bit & 7.99 & +51.17\% & 100.6\\
GLM: 2-bit, blockwise decode & 7.09 & +117.10\% & --\\
GLM: AutoRound/RTN & 7.83 & +94.27\% & --\\
\bottomrule
\end{tabularx}
\caption{Pilot Qwen3.8 27B compression.
PPL increase uses held-out C4 and BF16 PPL 9.9227. TPS is batch-one
decode speed ($127/L$ for the 128-token first-to-last interval $L$). GLM, Sol, Qwen, and MiMo denote GLM-5.3-Flash, GPT-6.1 Sol,
Qwen3.8-Flash, and MiMo-V2.6-Pro, respectively.}
\label{tab:calib-limits}

\par}
\end{table}

At 16 GiB, finer groups lower PPL but increase metadata storage;
the group-32 configuration exceeds the limit. At 12 GiB, mixed
three- and four-bit weights improve on uniform three-bit quantization.
At 8 GiB, similar package sizes yield different PPL. These results
motivate paired limits that expose precision-allocation and
representation decisions.

\noindent\begin{minipage}{\columnwidth}
{
\centering
\footnotesize
\setlength{\tabcolsep}{3pt}
\begin{tabular*}{\columnwidth}{@{\extracolsep{\fill}}lrrrr@{}}
\toprule
\textbf{Release} & \textbf{Package} & \textbf{Resident} & $\boldsymbol{\Delta}$\textbf{PPL} & \textbf{TPS}$\uparrow$\\
 & (GiB) & (GiB) & (\%) & (bsz=1)\\
\midrule
Official FP8 & 28.76 & 27.67 & +0.50 & 89.3\\
RedHatAI INT4 & 18.13 & \textemdash & +3.25 & 105.0\\
\addlinespace[3pt]
\rowcolor{RefFill}\multicolumn{5}{@{}l@{}}{\strut\quad\textit{Low-bit releases}}\\
sdkyuan QAT Q2\_0 & 9.04 & \textemdash & +22.05 & 101.4\\
ISTA-DASLab IQ2\_XS & 8.72 & 8.77 & +17.05 & 55.0\\
Unsloth UD-IQ2\_S & 8.68 & 8.57 & +13.83 & 55.2\\
Unsloth UD-IQ2\_XXS & 7.65 & 7.41 & +25.26 & 56.6\\
\bottomrule
\end{tabular*}
\captionof{table}{Released Qwen3.8 27B quantizations under the pilot protocol.
$\Delta$PPL is the percentage increase over BF16 (9.9227) on held-out
C4. TPS is batch-one decode speed, $127/L$, with $L$ the 128-token
first-to-last interval.}
\label{tab:calib-releases}

\par}
\end{minipage}

The pilot QAT Q2\_0 release uses llama.cpp; the remaining releases in
\Cref{tab:calib-releases} use vLLM.

A QTIP pilot package occupies 6.96 GiB with PPL 20.84\% above BF16,
but retains BF16 weights during execution; agent pilots also retain
high-precision copies. The low-bit GGUF releases keep resident storage
within 0.24 GiB of package size. These measurements motivate separate
allowances for saved packages and resident weights.

The released quantizations used in the main study are identified with
their verification results in \Cref{tab:artifact-measurements}.

\keepspace{6\baselineskip}
\section{Additional Experiments and Decision Analysis}
\label[appendix]{sec:additional-studies}

\subsection{Repeated Agentic Design}
We repeat GLM 5.3 and DeepSeek V4.1 Flash on one dense and one MoE
task (\Cref{tab:repeat-trials}). Four fresh trials use empty workspaces,
the original checkpoints, prompts, and verification protocol, Claude
Code through Fireworks, maximum reasoning effort, eight hours, and one
H200 NVL.

\noindent\begin{minipage}{\columnwidth}
{
\centering\footnotesize
\begin{tabular}{@{}llrrr@{}}
\toprule
Agent & Trial & PPL & HE/TQ/GS & Mean\\
\midrule
\multicolumn{5}{l}{\textit{Qwen3.8 27B / 12 GiB}}\\
DeepSeek & Orig. & 3.399 & 74/56/86 & 72.00\\
 & Repeat & 3.392 & 68/56/88 & 70.67\\
GLM & Orig. & 3.383 & 82/64/96 & 80.67\\
 & Repeat & 3.356 & 70/56/80 & 68.67\\
\midrule
\multicolumn{5}{l}{\textit{Qwen3.5 122B-A10B / 48 GiB}}\\
DeepSeek & Orig. & 2.803 & 82/76/96 & 84.67\\
 & Repeat & 2.820 & 86/72/96 & 84.67\\
GLM & Orig. & 2.814 & 78/74/96 & 82.67\\
 & Repeat & 2.819 & 82/74/96 & 84.00\\
\bottomrule
\end{tabular}
\captionof{table}{Original and repeated eight-hour designs on one H200 NVL.
Final verification measures perplexity (PPL) and accuracy (\%).
HE/TQ/GS denote HumanEval/TriviaQA/GSM8K, with fifty questions each;
Mean averages their accuracies.}
\label{tab:repeat-trials}
\par}
\end{minipage}

All four repeats pass final verification. Their PPL values lie within
0.81\% of the corresponding original designs, with the largest shift
in GLM's dense-model repeat.

\subsection{Controlled Quantization Ablations}
\label[appendix]{sec:paired-ablations}
We compare clipping-range selection and GPTQ error compensation in
GLM's repeated dense and MoE designs (\Cref{tab:dense-ablation}). Paired
variants share source weights, calibration, precision allocation, group
sizes, and inference. The MoE scales and channel order remain fixed.

\noindent\begin{minipage}{\columnwidth}
\centering\small
\begin{tabular}{@{}lrrrrr@{}}
\toprule
Setting & PPL & GSM8K & HE & TQ & Mean\\
\midrule
\multicolumn{6}{l}{\textit{Qwen3.8 27B / 12 GiB: three-bit MLP clipping}}\\
MSE search & 3.35560 & 80 & 70 & 56 & 68.67\\
Min--max & 3.36931 & 72 & 60 & 54 & 62.00\\
\midrule
\multicolumn{6}{l}{\textit{Qwen3.5 122B-A10B / 48 GiB: compensation}}\\
GPTQ & 2.81924 & 96 & 82 & 74 & 84.00\\
No compensation & 2.83444 & 96 & 76 & 78 & 83.33\\
\bottomrule
\end{tabular}
\captionof{table}{Quantization ablations of GLM's repeated designs.
PPL and accuracies (\%) use final verification. HumanEval (HE),
TriviaQA (TQ), and GSM8K each use fifty questions; Mean averages their
accuracies. Paired variants have
identical package/resident storage: 11.843/11.859 GiB (dense) and
47.835/47.854 GiB (MoE). The MoE control retains the selected scales
while removing GPTQ error~compensation.}
\label{tab:dense-ablation}
\end{minipage}

Clipping-range refinement in the dense model and GPTQ error
compensation in the MoE model both lower verification PPL at identical
model storage. Compensation reduces negative log-likelihood on 197 of
256 MoE verification contexts.

\subsection{Agent Responses to Feedback}
\begin{figure}[htb]
\centering
\includegraphics[width=\textwidth]{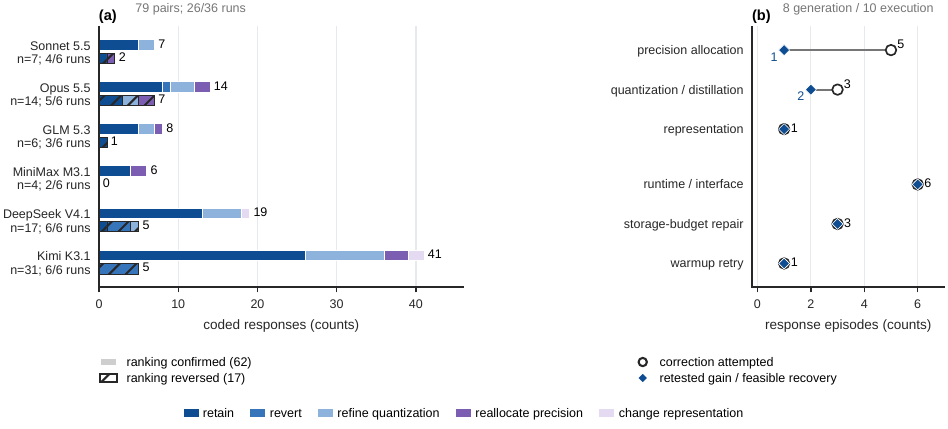}
\caption{Agent responses during compression design.
(a) Actions following 79 comparisons of local rankings with reward
feedback in 26 of 36 runs;
hatched bars denote ranking reversals, unhatched bars confirmations.
Actions may co-occur. (b) Circles indicate attempted repairs; diamonds
indicate matched local generation gains (eight cases, upper three rows)
or execution within both storage limits (ten episodes, lower three rows).}
\label{fig:agent-responses}
\end{figure}

\label[appendix]{sec:agent-response-analysis}
\Cref{fig:agent-responses} summarizes actions after local rankings and
reward feedback, together with subsequent generation and execution tests.

Six of eight local generation regressions prompt revisions. Opus and
Kimi improve on matched local tests while remaining below their
references. Ten execution or storage episodes later complete within
both limits, including one warmup retry.

\subsection{Resident Storage and Peak Device Memory}
For both releases, most additional device memory is allocated during
loading, with smaller increases during prefill and decode
(\Cref{tab:peak-memory}). Runtime allocations include caches,
recurrent states, and temporary workspace.

\noindent\begin{minipage}{\columnwidth}
\centering\small
\begin{tabular}{@{}lrrrr@{}}
\toprule
 & Resident & \multicolumn{3}{c}{Sampled device peak}\\
Model & storage & Load & Prefill & Decode\\
\midrule
Qwen3.8 & 10.953 & 13.000 & 13.349 & 13.353\\
Qwen3.5 & 43.349 & 45.332 & 45.506 & 45.510\\
\bottomrule
\end{tabular}
\captionof{table}{Resident storage and device peaks (GiB) for Qwen3.8 27B GSQ-RCO
IQ3\_S and Qwen3.5 122B-A10B Unsloth UD-IQ3\_S. Values are medians of three cold starts.
Device usage is sampled every 10 ms at batch one with 2,048 input and
128 output tokens and a 4,096-token single-sequence cache.}
\label{tab:peak-memory}
\end{minipage}

\end{document}